# Control and Automation for Industrial Production Storage Zone: Generation of Optimal Route Using Image Processing.

Huerfano Z. B. A., Jiménez F.
2017, ba.huerfano10@uniandes.edu.co, fjimenez@uniandes.edu.co

*Abstract* – Digital image processing (DIP) is of great importance in validating and guaranteeing parameters that ensure the quality of mass-produced products. [1]. Therefore, this article focused on developing an industrial automation method for a zone of a production line model using the DIP. The neo-cascade methodology employed allowed for defining each of the stages in an adequate way, ensuring the inclusion of the relevant methods for its development, which finally incurred in the modeling, design, implementation, and testing of an optimal route generation system for a warehouse area, using DIP with optimization guidelines, in conjunction with an embedded platform and the connection to programmable logic controllers (PLCs) for its execution. The system was based on the OpenCV library. [2]; tool focused on artificial vision, which was implemented on an object-oriented programming (OOP) platform based on Java language. [3]. It generated the optimal route for the automation of processes in a scale warehouse area, using the segmentation of objects and the optimization of flow in networks as pillars, ending with the connection to PLCs as a method of action, which in case of implementation would eliminate constraints such as process inefficiency, the use of manpower to perform these tasks, inadequate use of resources, among others.

## I. INTRODUCTION

The electronic age and the development of automation have led to the creation of new products, as well as the growth of the industry. [4], This is due to the search for quality and cost reduction of the processes moving the industry towards the use of state-of-the-art technologies in order to achieve a cleaner, faster, and more accurate process.

The current technological development presents a constant evolution, as far as software development is concerned, as pointed out by Gonzales [5], This has brought with it numerous advantages, including the development of more efficient, faster and cheaper technologies [6]; because of the above, it is plausible to address certain issues that were ignored some time ago due to the requirement of large computational resources. This is due to the fact that new technologies have managed to improve information processing speeds, thus opening a wide path, such as the study in the development of DIP algorithms.

Therefore, the current document presents the development of a project focused on the control and automation of a zone of an industrial production line model, using the DIP for the segmentation, identification, measurement and transfer of objects within a scaled environment, using a platform based on a high-level object-oriented language. In order to innovate in the area of image processing and bring benefits to the Colombian industry. This project was oriented to a research point of view, of exploratory type.

During the development of the project, the methodology used was the application of a neo-cascade model [7], which allowed the development of tasks in a sequential manner, allowing the review of each of its phases, as follows.

The requirements analysis identified the DIP technology trend in industrial control, abstracted the characteristics of the industry, defined the production line model and chose the development platforms. The system design used what was identified in the previous phase to define the area of the production line that was the focus of the project, defining also the automation to be applied, as well as the control scheme applied and the methods used to fulfill it. The implementation phase carried out what was proposed in the design, through the creation and adaptation of selected methods and platforms. The final testing phase shows the results and changes obtained from the final implementation of the system, showing the execution speed, efficiency and scalability of the system.

## II. REQUIREMENTS ANALYSIS

This section contains the elements used for the acquisition and identification of the requirements demanded by the project, in order to identify the benefits, disadvantages and basic characteristics of the technologies implemented, thus providing a clearer and more sustainable context for the development of the project and its subsequent phases.

### A. Bibliographic Review

A literature review was developed in order to identify the scientific context related to industrial control through the IMP, in order to directly define the models, strategies, techniques and technologies that have been involved in the context of control, through the IMP worldwide, allowing to have a basis to capture and beneficially apply the identified values to the development of the project.

For the purposes of a good approach to the bibliographic exercise, guidelines and characteristics were defined and taken into account for the selection of information such as: Chronological search window (**15 years**)**,** Minimum number (**25 references**), Geographical framework (**Worldwide**)**,** type of content (**Scientific - Focused on control and automation by means of the DIP**). In addition, basic characteristics were defined for the selection of the most valuable documents for the bibliographic search exercise, using the items shown in *Table 1*.

*Table 1*. Criterios de Búsqueda.

| Search tools | Type of documents | Chronological Range |
|---|---|---|
| Databases | Articles | |
| Virtual Libraries | Books | |
| Indexed Journals | Web Sites | 2001 - 2016 |
| Search engines | Journals | |
| Metasearch engines | Projects | |

Where the documents found were related bibliographic files, with a total number of 39 documents studied of which, through this exercise identified a number of 27 articles focused on linking control and automation through the IMP, and the remaining 12 focused on quality control. A positive result was obtained, as far as the total number of documents is concerned, with a value of 14 additional documents.

Analyzing the features of the abstracted development of this exercise, it was identified that there is currently low level of progress of type applied to the industry, as few of the related documents focused on industrial level and most of the control and automation in an investigative way. It is worth mentioning that the existence of a tendency to process control through the DIP was identified, likewise quality control is a topic addressed in a large number of documents, but no direct applications to industry were found, but only abstractions of problems and design of solutions. Finally, the exercise allowed us to identify the feasibility of using this type of technology for process and quality control. And it helped to verify the methods currently used for such tasks.

*B. Requirements Gathering*

The collection of pertinent information was developed for the subsequent design of the automation and control system for the industrial production line area; thus, determining the particularities of the project and selecting the different characteristics that later became part of the basis for the development of the platform.

*1) Definition of the Productive Sector - GDP Study.*

The productive sector was defined in order to identify and address the characteristics immersed within the line of production that was subsequently selected, this was done from the data provided by the Ministry of Commerce, Industry and Tourism through monthly and annual studies delivered by this entity, where the participation of each of the sectors within the Gross Domestic Product (GDP) is defined, as well as its growth and participation for the department of Cundinamarca; it is worth clarifying that this department is selected because it is our social and cultural environment.

*Table 2. Characteristics of the death of productive sector.*

| Used Studies | Selection aspects |
| --- | --- |
| • Industry reports January to December 2015.<br>• Industry Report January 2016.<br>• Economic and Commercial Profiles (National, Cundinamarca and Central Eastern Region). | • GDP participation.<br>• Growth of the sector.<br>• Contribution to the growth of the industrial sector.<br>• Competitiveness. |

The analysis showed that the most dynamic sector, which complies with the aspects indicated above to a greater extent, was the **beverage** manufacturing sector, based on the fact that:

- It had a growth of in real production and a contribution of percentage points to manufacturing industry growth for January 2016[8].
- It was the most dynamic subsector in the last 12 months as of January 2016, contributing percentage points [9, 10].
- The good performance of this subsector was due to higher orders from large supermarkets and chain stores, advertising campaigns and marketing strategies to increase their customers, both in commercial and institutional consumption channels, the launching of new presentations and references according to the following factors [8].

*2) Identification of industries in the sector.*

A direct survey was carried out within the Colombian industries with the greatest national and Cundinamarca contributions in the beverage sector, where a clear advantage of the companies Bavaria S.A. and Postobón S.A. was easily identified. The selection of Colombian companies was due to the geographical framework of the current project and was based on the data identified in [11, 12, 13].

It can be seen that they have different market segments and products, but these companies have similar production lines for returnable products. And it is there where the topics of the following numerals and their characterization are focused.

*3) Production line model*

In order to identify which parts of the production chain are immersed in the chosen companies, it was necessary to define the production line model on which we worked, and for this purpose we made use of what was exposed in [14], where they express the model usually used in brewing production companies.

It was identified within the model which tasks of the process are observable by the DIP; allowing in turn, to know what should be the base production line on which the project worked. The production line zones identified were packaging, packing and storage [15, 16], ] which have defined processes, as illustrated in *Figure 1*.

The zones are delimited with red, green, and blue colors assigned respectively to the packaging, labeling and warehousing zones. In subsequent paragraphs, the selection of the zone of the production line to which the automation of its processes was applied was carried out.

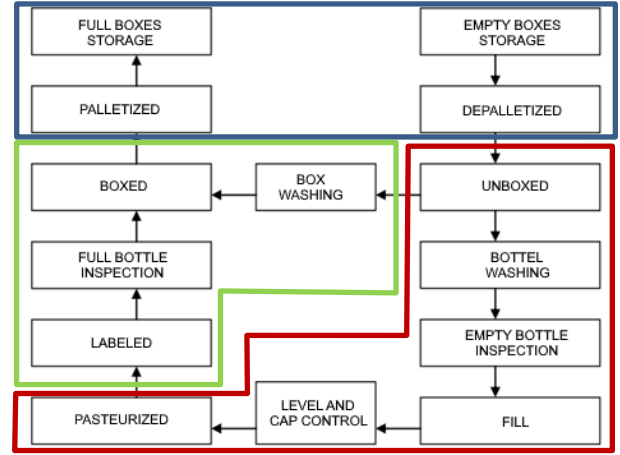

*Figure 1. Phases within the production line[1].*

*C. Development Software Selection*

In order to identify the programming language, platform and development environment to be used to successfully complete the work, an analysis of three OOP languages was carried out, which currently have the necessary characteristics for DIP development (Java, C# .NET and M) and for usability reasons it was defined that they should have a graphical interface that would allow their adequate use; the languages are described in the following *Table 3*.

*Table 3. Properties of programming languages.*

| *Language*: | *Java* | *C# y .NET* | *M* |
| --- | --- | --- | --- |
| *Platforms* | Android, IKVM.NET, Development Kit, EE, SE, OpenJDK … | Windows phone, OS Windows, Silverlight, WCF, WPF, Azure... | MCR, C, FORTRAN. |
| *Development environments* | Eclipse, BlueJ, JBuilder, Jcreator, Netbeans | Visual Studio, MonoDevelop | Matlab |
| *Executable* | Yes | Yes | No |
| *Image processing* | OpenCV, JavaCV, JWT, SIWING | EmguCV, OpenCV | Image Processing Toolbox, OpenCV |
| *Licensing* | GNU General Public License | Payment (City, Professional and Company) | Payment (Standard, Educational, Home, ...) |

*Sources: Learn more about Java technology [3], What can you do with visual estudio? [17] and MATLAB Key Features [18].*

Considering these specifications, any of the languages studied is capable of developing the current proposal, but for reasons of versatility, number of libraries for image processing, licensing, and development environments, it was decided to select the JAVA language as the basis for the software development necessary for the implementation and testing of the project.

The development environment used was Netbeans IDE 8.1 (*Latest stable version available since November 2015*) because it has a programmer-friendly graphical interface, the possibility of using modules created by a worldwide network of programmers and a history since 1999 [19].

---

[1] Source: (Study of a packaging line and application of the TPM Methodology to increase its efficiency, through the reduction of small stops in a packaging grouping equipment[16] – edition by autors)

Finally, the selcted library was Open Source computer vision OpenCV [2], due to its support in multiple programming platforms and the advantages that a post-development code migration would entail, bringing with it also facilities in the search of information related to this library.

### III. SYSTEM DESIGN

The identification and design of different parts of the project are immersed in the current section, based on what was abstracted in the immediately preceding stage (II), the necessary constructs for the system were completed, which allowed not only to have a basis for the development, but also to have an idea of what was later brought to fruition in the implementation phase (IV).

#### A. Production Processes

The processes involved in each zone of the production line were identified and described in order to find out what their characteristics were [15, 16, 20, 21, 22], where the beer production processes are described, immersed in the phases shown in *Figure 1*.

From these phases, the processes were identified, allowing to characterize the operations that must be controlled in them. It was defined which of the three zones represents the greatest gain for the production line, from the point of view of automation with DIP.

*1) Production quantity*

This characterization was of vital importance because it provided important characteristics (sampling rate per process, operation speed, installed capacity...). The amount of production is inherent to the process and varies depending on the company, therefore, it was evident the extraction of data concerning production lines referring to Bavaria S.A. at Colombian level, where it was identified that it is the largest brewery in Colombia, [23] one of the largest in America and the tenth largest in the world (now part of SABMiller) [24].

According to data provided by works [25, 26], it was possible to identify the production capacities of Bavaria S.A.'s plants, resulting in minimum and maximum capacity data about their brewing production lines: production between $72.000\ to\ 554.000\ Bottles/Hour$ and storage around $1'500.000\ to\ 10'000.000$ bottle tens.

This data corresponds to the number of bottles manufactured per hour of the most productive lines at each site, as well as their storage capacity. Now taking as a premise the strongest case, it was identified with a sampling rate of 154 frames per second ($\mathbf{154\ fps}$), which would correspond to capture individually the bottles on the process, and in the storage the administration to $\mathbf{10 millions}$ of tens.

Finally, it was decided that the area in which the project would be developed should be one where the effect of automation is on a large scale, so it was decided to develop the *warehouse area*, as it does not have the full automation of its processes according to [15, 21, 22], where personnel must perform tasks individually in order to fulfill and supply the operation of the area. On the other hand, it should be noted that this process benefits the IMP to a great extent because its sampling level is not so high compared to other stages in other zones.

#### B. Warehousing Area Automation

Ahora se definió cómo abarcar el modelo de control expuesto más adelante en III.C, for this purpose it was identified in [15, 16, 20, 21, 22], that at the warehousing level the processes are carried out by personnel, who are in charge of moving the elements from one area to another, using forklifts and keeping control of the goods; in addition, due to safety regulations, exclusive lanes are used for the transport of goods; this safety is imposed due to the interaction of human resources, which directly affects the execution time of the tasks and produces errors in the development of the work.

It is there where the current project developed its core, in order to eliminate that human barrier, which restricted the possibility of a better use of resources and greater efficiency in the development of the process; The project broke the scheme of the use of security zones in the transportation and handling of stock, where it was defined that the best option was to have the optimal route for the transfer of the elements, which would greatly help to improve the process there, in addition to saving resources in hiring personnel, execution time and prevention of failures.

This abstraction was possible to carry out in this way, due to the following characteristics that help its approach.

- *Start position*: It was identified that the use of a forklift allowed to directly abstract the starting position for the desired control and therefore allowed to easily identify its start within the algorithm. It was also noted that such a forklift could be controlled by PLC, thus representing versatility of motion control.
- *Characteristic colors:* The warehousing at the industry level, has identifiable characteristics by color, such as the truck unloading area, which usually has a distinctive color, the objects and merchandise arranged there have unique colors compared to their surroundings, the forklift has its own color and additional users immersed in the warehouse have uniforms and helmets that easily identify them by color.
- *Elimination of exclusive lanes:* The security scheme that depended on the users was eliminated, which through the use of exclusive lanes performed the task; therefore, now they were free to use empty spaces within the warehouse for their movement.
- *Spatial location in 2D:* We used spatial location in two dimensions only, because we know the height of the elements immersed in the system, thus eliminating one of the dimensions, allowing a more appropriate and effective application of control, also this type of view allowed the use of Cartesian mathematical tools, which employ little computational level and the use of simple geographic positioning.

#### C. Basic Control Scheme

Once the phase of the production line on which the project was focused (***warehousing***) was defined, we proceeded to define the control scheme with which it was approached, where we opted for a *distributed control* system *with data centralization*, so that the automated area would have autonomy in the application of control and in turn allow coordinating larger actions or changes at a global level thanks to the centralization of data.

This type of control was defined by modules, which have autonomy for the development of automation and control activities in the warehousing area, but can be subordinated to orders created from the centralization of data, which allowed control to be carried out at various levels and thus assign different hierarchies of system administration.

However, each module had to have a minimum level of information processing, means of communication and a method of acting on the system. Therefore, the modules were composed by the participation of an embedded system, which provided the necessary level of processing and the connection of a control item performed by a PLC, therefore, each control module is composed of a processing part (embedded platform) and an actuation part (PLC).

*1) Embedded system selection*

In order to select the most suitable embedded system, it was necessary to carry out a review of such systems worldwide, in order to identify the cards that, to date, had the best features and the best performance for the project.

For this purpose, use was made of comparison portals for this type of equipment, such as OpenBenchmarking.org [27] and socialcompare.com [28], where the data found were practically identical, having only variations with respect to prices and weights of the cards, the data were verified with the official websites [29, 30, 31, 32, 33] of each item compared. The aspects evaluated are shown in the following *Table 4*, 4where the differences between each of the systems can be observed.

It was observed first hand that the last two cards did not have OpenCV support, previously selected image processing libraries (II.C) were therefore directly discarded. Attention was therefore focused on the remaining three elements.

The level of documentation and indexing of these platforms on the web, which is an indicator of high reliability in terms of support and project development, was used as decisive data; therefore, the Google Trends tool (Documentation and web indexing - ODROID, Orange Pi and Raspberry Pi ) was used to verify the argument.

*Table 4. Comparison between embedded systems - February 2016*

| Technical Details | Raspberry Pi 3 | Orange Pi Plus | ODROID C1 | Pine A64 | C.H.I.P. |
|---|---|---|---|---|---|
| CPU | 1.2GHz 64-bit Quad-core ARMv8 | 1.6 GHz 32-bit Quad-core Cortex-A7 | 1.5GHz 32-bit Quad-core Cortex-A5 | 1.2GHz 64-bit Quad-core ARM A53 | 1 GHz Allwinner A13 Compatible SoC |
| GPU | Broadcom VideoCore IV @ 300MHz | ARM Mali-400 MP2 @ 600MHz | ARM Mali 450 MP2 | ARM Mali-400 MP2 | ARM Mali-400 |
| RAM | 1GB | 1GB | 1GB | 512MB | 512MB |
| LAN | ✓ 10/100 | ✓ 10/100/1000 | ✓ 10/100/1000 | ✓ 10/100 | ✗ |
| Wi-Fi | ✓ | ✓ | ✗ | ✗ | ✓ |
| HDMI | ✓ | ✓ | ✓ | ✓ | ✗ |
| Bluetooth | ✓ | ✗ | ✗ | ✗ | ✓ |
| USB | ✓ 4 | ✓ 4 | ✓ 4 | ✓ 2 | ✗ |
| Weight | 45g | 50g | 40g | 46g | 26g |
| Price | USD$ 30 | USD$ 49 | USD$ 38 | USD$ 15 | USD$ 9 |
| Operating Systems | 21 | 13 | 14 | 26 | - |
| Camera | 30 fps 8Mpx | 15fps 5,1Mpx | USB 25fps VGA | 30fps 5Mpx | 30fps 0,3Mpx |
| JAVA | ✓ | ✓ | ✓ | ✓ | - |
| OpenCV | ✓ | ✓ | ✓ | ✗ | ✗ |

Source: *Open Benchmarking [27], Social Compare [28], Raspberry Pi 3 Model B [29], Orange Pi Plus [30], ODROID-C1 [31], Pine A64 [32] and C.H.I.P. [33].*

Finally, it was decided to use the **Raspberry Pi 3 (RPi 3) - Model B** platform, due to the results obtained in the two previous analyses, where this platform stood out for its number of operating systems, more efficient camera module, greater number of necessary technical details and price; in addition to a greater relevance and participation in terms of web documentation, which in comparison to its competitors is abysmal.

*2) Peripheral and operating system selection*

Once we defined the embedded platform to be used (RPi 3), it was necessary to identify the peripherals and the operating system (OS) to be used.

- *Micro SD (uSD):* the RPi 3 makes use of this technology to mount the file system and the OS image. it is necessary to identify a uSD memory with good performance, in order to obtain the highest block transfer speed, reading and writing of information, as this directly influences the execution speed at which the RPi 3 can work**;** Reason to use an uSD memory with *Ultra High Speed 2 (UHS II)* with $32GB$ storage capacity, which far exceeds the minimum recommended in the official RPi 3 documentation [34] ($8GB\ Class\ 6$).
- *Camera module:* Making use of the information captured in the **Table 4** which was verified with [29] The official RPi module known as **PI NOIR CAMERA V2.1** at commercial level was used; this module makes use of a Sony IMX219 8-MPX camera [35], providing up to 30fps at its highest resolution. This module was selected because it uses the CSI (Camera Serial Interface) port, which allows fast data transfer.
- *Protection and cooling system:* A 7-layer protector was acquired, as well as heatsinks for the most thermally stressed components (Processor, GPU and RAM) and finally a $5VDC$ fan for the cooling system, in order to avoid temperature problems in the development of project tests, since this requires large resources.
- *Operating System:* The OS was chosen according to the recommendations given in [34], where the use of Rasbian (Debian based distribution) is recommended, therefore the latest Jessie Lite version (2016-11-25) was used, the lightest version of the OS that does not include a graphic interface was chosen, because all the drivers and programs necessary during the course of the project were adapted, thus avoiding the use of resources both in RAM and in the uSD, allowing to have the features clearly necessary for the application.

*3) PLC connection*

As indicated above, the control module was defined as the element that has processing (RPi 3) and actuation, so it was decided to make use of the connection to a PLC network, this because the current industry makes use of this type of equipment for the control and automation of processes as evidenced in [15, 16, 22], Therefore, it was of vital importance to take advantage of the infrastructure found in the project environment.

With this in mind, we made use of the elements provided by the Universidad de los Andes in the Electronics laboratories, where we have access to Siemens PLCs with S7-300 references [36] y S7-1500 [37], which provided an evaluation space for connection and reception of information necessary for the control. Once the connection of the data blocks was achieved through the modification of variables, the transfer of information that was necessary for the PLC to perform its task was achieved as a consequence.

D. *Segmentation Algorithm Design*

In the current issue, the detection and segmentation of colored objects in an environment composed of different objects is the essential problem; for this purpose, a pixel-by-pixel decomposition of the image was used. The algorithm is able to identify the distance and the center of the object (of a specific color) with respect to the other objects and thus show the results obtained from the image, in addition to segment and identify each object, regardless of whether they have the same color.

In pursuit of this objective, general aspects to be addressed were identified and defined in order to implement a robust and complete algorithm.

*1) Characterization of the work environment*

We proceeded to identify the basic characteristics of the study environment, which will be framed with the main features of a real automated industrial work area [6], where this area is characterized by cleanliness, non-intersection of different work areas, environment conducive to control, among others.

A monochromatic background that allows the identification of objects, defined with white color, was defined as indispensable characteristics for a favorable environment, the implemented background has $4m^2$ area on the testing setup. And the location of the camera in the frontal plane in order to achieve a general mapping of the background used in 2D

The environment had good lighting conditions and front lighting was implemented, which allows uniform illumination and is commonly used in object detection systems [38].

*2) Acquisition of images*

Data acquisition was performed through a camera (Logitech HD Pro Webcam C920) via USB 2.0 protocol: this was used specifically in the context of the design of the segmentation algorithm, thus providing good image quality and efficiency in the transmission of information[39].

Which was located to $208cm$ in order to allow a pixel for the resolution imposed, represent a distance $2.96875mm$ by side, this measure was defined by the camera coverage area with $190cm$ length and $142.5cm$ wide, to the total $2.7075m^2$.

*3) Color scale selection*

The selection of the scale was based on the acquisition of information from the camera, which identified the use of the primary colors in this case of the RGB scale (Red, Green and Blue) since the coding model used by the camera in question is the mentioned (RGB), in order to avoid losses of computational resources and efficiency of the algorithm in the transformation to another type of scale.

*4) Sampling time*

The sampling time as a goal defined for the current design was defined in search of having a minimum continuity of frames, close to the imperceptible by the human eye, on the other hand, the amount of samples per unit of time is limited by the maximum possible amount of the camera used, being a maximum of $30 fps$. Finally, taking into account a prudent amount of processing time of half of the capture time, a target margin of minimum $15 fps$.

*5) Characterization of objects*

The objects immersed in the environment had distinguishable characteristics at first glance, which allowed us to read the results more easily in the current document. Therefore, we defined the colors Green, Red, Blue and black, these colors will be the ones that will be segmented in our work environment. Additionally, we defined a minimum size of $15 cm^2$ for any object immersed in the environment, in order to avoid that the appearance of noise immersed in the image could affect the measurements obtained.

*6) Identification by color*

To detect an object of a certain color, it was initially necessary to identify all the pixels that compose it. To do this, a search was performed on all the image data, calculating whether or not they belong to the desired color. In order to differentiate each color evaluated at that moment, first the selection of characteristic data containing the color pixel and finally the proximity to the desired color is performed, in order to classify it or not within the type of color sought, by means of the super position of a range.

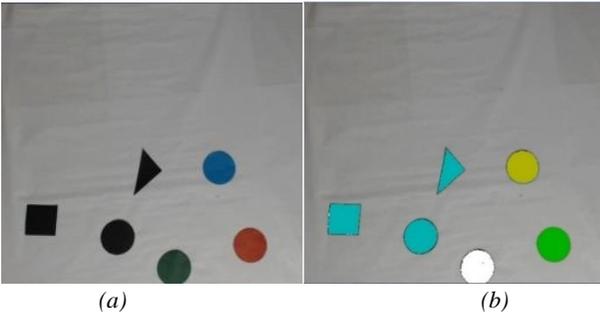

*Figure 2. (a) Captured image and (b) Color processed image.*

*7) Positioning*

To locate the objects and the desired color pixels, we saved the coordinates of each pixel identified and classified in the desired colors (Red, Green, Blue and Black) thanks to the sum of their positions, we calculated the midpoint of our object through the following formulas:

$$\frac{\sum_{i\in OD}^{n} i}{\sum_{i\in OD}^{n} 1} \quad y \quad \frac{\sum_{j\in OD}^{m} j}{\sum_{i\in OD}^{m} 1} \quad (1)$$

Where $n$ and $m$ represented number of pixels on length and width image size. $OD$ as the set of pixels of a desired color object. $i$ is the integer value of the position $x$ y $j$ the integer of the position $y$ of a specific pixel.

These were used to obtain the mean, which represented the coordinate $(x,y)$ of the pixel of the center of the evaluated object, which later served as a basis for the calculation of the distance between these objects and also for their segmentation. These coordinates of the midpoint will be quite accurate if we have a large number of pixels of the object, and will be very useful to locate the object with respect to our robotic platform.

Once we have the coordinates we can infer whether the object is to our left, right, above or below our moving platform or the arrival point (red colored area). We proceeded to find the distance of the moving platform with respect to the arrival zone (in pixels) using the hypotenuse equation:

$$Distance = \sqrt{(x_2 - x_1)^2 + (y_2 - y_1)^2} \quad (2)$$

Where $x_k$ represents the coordinate $x$ and $y_k$ the coordinate in $y$ for a given pixels of the object $k$, obtaining the position and distance between objects, it is known that the image area, indicating the approximate real area occupied by the actual object like ($\sum_{i\in OD}^{n}(2.96875mm)^2$). But due to the nature of the irregular objects it became necessary to frame the object in a slightly wider square area to have a margin of distance between an obstacle and what was defined as a moving platform for the document, for the purposes of the study was defined the color black as objects, to achieve this purpose was obtained the minimum and maximum coordinate of the desired obstacle:

$$\left(\min_{i\in OD} i, \min_{j\in OD} j\right) y \left(\max_{i\in OD} i, \max_{j\in OD} j\right) \quad (3)$$

Thus, providing a square area that frames the obstacle under study.

*8) Segmentation*

The segmentation of objects consists of the division or separation of objects from the environment, in order to indicate that they are different and that they are in another part of the plane to the one being observed. In addition, it allows to have a perspective of the number of objects being observed and also to generate a positioning of each one of them with respect to the moving platform[40].

The segmentation of the objects was done by means of the difference between the distances of the agglomerations of pixels belonging to the objects, for which we made use of a defined range of $10\ px$, where a difference between the last pixel $\in OD$ in the image with respect to the next one in the same row or column, will represent a new possible object, if the new agglomeration of pixels found exceeds the area mentioned in III.D.5), is considered as a new object, different from the first one found.

This way, multiple objects of the same color can be detected, which, together with the characteristics expressed in III.D.4) allowed its total segmentation. In this case it will be the objects defined as possible obstacles (black color).

Sabemos que hay distintos objetos que identificar, y que unos We know that there are different objects to identify, and that some pixels will belong to a possible obstacle, other pixels will belong to others. How do we know which object it is? To do this we will use an instance of the OOP we are using, which allowed us to create entities known as objects, allowing us to store the information of any possible obstacle present in the study area. Finally obtaining what is represented by the *Figure 3*.

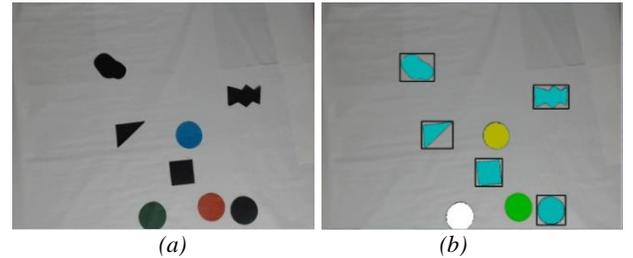

*Figure 3. (a) Captured image and (b) Segmented image.*

Where it can be seen how the aforementioned characteristics are reflected in *Figure 3 b)*, where possible obstacles regardless of their shape are framed and identified. Without being affected by the proximity with another object of different color, which are also still fully positioned and located.

*E. Optimal Path Algorithm Design*

Here we present the design of the algorithm that supported the creation of the optimal route in the implementation phase, thus presenting two possible options for the generation of routes, from which the most efficient one was subsequently selected. This was done in order not to propose only one solution, but to address the issue from several approaches. The current section only shows the selected method, in order to avoid extension.

*1) Optimization in control*

At the level of control through the IMP, there is the participation of different variables, which are dependent on the characteristics and environments of the agents within the control of a specific problem or approach; such controls are focused on solving problems such as service optimization, increasing the reliability index, reducing mechanical stress, process automation and reducing the cost of production [41].

Optimization has allowed to generate advances in such controls through algorithms that allow to identify the optimal way to use resources and strategies for the development of a control and that under open or closed circumstances allow to optimize the operation of the control [42]. Specifically speaking we can define optimization as the methods to determine the best solution to a mathematically defined problem, which usually models a physical phenomenon this as stated by Fletcher [43] for this we will evaluate some methodologies implemented in this subject.

a) Linear optimization.

It is based on solving problems with linear type equations or linear combinations which are governed by different constraints of the same type, such problems can usually be abstracted directly from their behavior or from their requirements [44]. For this type of optimization there are different solution methods such as: Simplex Method, Revised Simplex, Complementary Duality, Ellipsoid Method (Interior Point), Minimum Cost Flow, Dijkstra's Algorithm and Maximum Flow. Of which most of them are modifications and variations of the initial simplex method [45, 46] which was presented by George Dantzig in 1947.

Once these concepts were identified, the most appropriate method for solving the routing problem was selected, which was Dijkstra's Algorithm.

*2) Problem characterization*

The problem addressed in the current paper was referred to the selection of the optimal route of a mobile robotic platform within a defined environment for the transfer from a specific point to another of the study environment, for such purpose the development was based on the DIP.

In order to be able to make the most appropriate identification of the nature of the problem, were implemented a detection algorithm, segmentation and location of objects within the study environment (III.D), thus obtaining an information base and thus achieve our problem of the environment.

a) Abstraction of the problem.

In this case it was necessary to identify the type of problem to be addressed, where due to its characteristics it was defined as linear, and taking into count the information obtained by the algorithm, an objective function is identified as follows.

$$\sum_{i \in G}^{m} \sum_{j \in G}^{m} x_{ij} c_{ij}, \qquad m = \max i \in G \qquad (4)$$

Where $G \in \mathbb{N}$ represents the set consisting of the robotic platform, obstacles and arrival zone, $m$ the number of elements that make up the set $G$ which are designated by a number from 1 to $m$, identifying the number 1 as the robotic platform and $m$ the arrival point, $x_{ij}$ e is the flow value allowed by the junction between the element $i$ to the $j$, and $c_{ij}$ the cost of using such a junction. All this was done based on the general theory of directed graphs [47] in a similar way to what has been exposed in different studies [48, 49, 50]. Taking consideration the above mentioned, we proceeded to generate the pertinent constraints according to the problem.

$$\sum_{j \in H}^{m} x_{1j} = 1 \qquad (5)$$

$$-\sum_{i \in I}^{m} x_{im} = -1 \qquad (6)$$

$$\sum_{(j \to i) \in L}^{m} x_{ji} + \sum_{(i \to j) \in L}^{m} x_{ij} = 0, \qquad i = 2,3,\dots,(m-1). \qquad (7)$$

$$1 \geq x_{ij} \geq 0 \qquad (8)$$

Where $H \subseteq G$ represents the node set with ouput flow from the source node with $c_{1j} < \infty$, which colloquially means that they have direct connection with the source node (mobile platform), $I \subseteq G$ and is the set of nodes with incoming flow to the sink node with $c_{im} < \infty$ this indicates the nodes with direct connection to the sink node (Arrival zone).

The constraint shown by equation (**5**) represents that the summation of all outflows from the source node (number 1), with the adjacent nodes belonging to $H$, will be equivalent to **1**, showing that only one flow path will assign to this value, allowing the existence of only one possibility of improving.

The **constraint** (**6**) represents that the summation of all incoming flows to the designated sink node (number $m$) with the adjacent nodes belonging to $I$ must be $-1$ indicating that only one node will be assigned with that value ensuring only one possibility of arrival.

The restriction (**7**) is based on the conservation theory of flow [51]indicating that all inflows and outflows of the nodes will be 0, except for the source and sink node which will have values of 1 and -1 respectively, where the set $L \subset G$ represents all nodes excluding the source and sink node ($L = \{2,3,\dots,(m-1)\}$).

Finely (**8**) indicates the possibility of the existence of a flow 1 or the absence of it 0.

b) Método Dijkstra

This method use the abstraction and features imputed above, the current method makes use of the architecture that provides the flow in networks to find the shortest path from one node to another.

The algorithm was initially exposed in 1959 by Edsger Dijkstra [52] This algorithm allows obtaining the shortest path from a node to all the nodes included in the network. The main idea is to mark all the shortest paths that start from the origin node and lead to the other nodes. Once the paths to all the nodes have been found, the algorithm ends; in our case it stops when it finds the shortest path between node **1** and the $m$. The algorithm be described by the following steps:

- Initialization of costs as infinite except for the origin node which will always have a value of 0, $c_{ij} = \infty \ \forall i,j \in G$.
- The current node is saved, and the next lowest cost directly connected node is searched for and visited.
- The visited node is marked with the cost from the origin and its immediately preceding node.
- We go through all the nodes adjacent to those already visited, depending on the value of the cost reflected from the origin node (add costs), we continue selecting the lowest one.
- Return to the second item in the list, until all nodes are marked as visited.

c) Final considerations.

What was identified in the previous paragraphs generated a very close abstraction of the real problem, but it was still necessary to consider the restrictions of the size of the obstacles and the free space between them. These restrictions were taken into account outside the Dijkstra algorithm, which were assigned to the object segmentation algorithm, where the identification of obstacles, and nodes belonging to $G$, in the case that the distance between the objects was too small to allow the robotic platform to pass between these obstacles, it was determined whether or not the distance was large enough to allow the robotic platform to pass between them. In case it was a very small one, it was taken as an object conformed by the two objects under evaluation and the space between them, besides having an additional preventive distance dictated by the coverage area of the obstacle.

IV. **IMPLEMENTATION**

Based on the constructs developed and the design proposed for the project, we proceeded to the implementation of each element, to make each one available and finally generate the total system of the project. Therefore, this section shows the development and adaptation of each part previously proposed, which will be united in the final testing phase.

## A. Embedded System - Raspberry Pi 3

For the development of the project, the implementation and commissioning of the chosen embedded system ($RPi\ 3$) had to be carried out, based on the requirements set out in III.C and also following the recommendations provided by the official RPi documentation [29, 34, 53].

### 1) Operating system installation - Raspbian

The OS selected in the design phase was installed, which corresponded to Raspbian in its latest version Wheezy until August 2016, on which all the adaptation was developed; it is worth mentioning that due to operating system upgrade in November 2016 the distribution was changed to Jessie [54].

For the installation we resorted to downloading the Raspbian Wheezy OS image, which came pre-compiled for use on the RPi; with this image we proceeded to deploy it on the chosen RPi. $uSD$ chosen in III.C.2) by giving the proper format ($FAT32$) to the memory, in order not to generate corruptions in the file system and its reproduction.

### 2) Driver update and installation s

Once the OS was functional on the RPi 3, all the information was updated in order to obtain the latest version of all its programs and drivers. This way we managed to have the OS and programs up to date, ensuring a more stable distribution.

Next, the drivers for the use of the PI NOIR CAMERA V2.1 camera module were installed, in order to be able to access its instance virtually, therefore, the OS configuration was modified and the CSI use was enabled, thus obtaining access to the camera module, then the necessary drivers were installed, and finally we added a new device to the OS, being this the connecting entity between the camera module and the OS. $modprobe\ bcm2835 - v4l2$ to the OS, being this the connecting entity between the camera module and the OS.

### 3) RPi 3 Overcloking

Although the selection of the embedded system was based on the fact of having good performance, it was additionally decided to make use of the overclock method (acceleration outside factory limits), in order to obtain an improvement in processing speed, for this purpose changes were used within the document that loads the OS when performing Booting, this document has the basic guidelines on which the RPi works, therefore if the limits are modified a little and making the appropriate adjustments it is possible to obtain a higher speed by the equipment, without it being affected. Finally, if the acceleration was carried out within the appropriate limits, which resulted in a small increase in temperature, which was mitigated by dissipation methods.

*Table 5. Overcloking applied to the Raspberry Pi 3.*

|  | **Normal** | **Obtenido** |
|---|---|---|
| **Procesador** | $1,2GHz$ | $1,4GHz$ |
| **GPU** | $64MB$ | $256MB$ |
| **RAM** | $900MB$ | $740MB$ |
| **Lectura / Escritura (uSD)** | $11Mbps / 19Mbps$ | $32Mbps / 70Mbps$ |
| **FPS** | $30.0$ | $90.0$ |
| **Tamaño imagen** | $640X480$ | $1024X768$ |

To avoid performing deterministic tests to find the values shown in the table, we made use of small studies carried out on the subject with the equipment (RPi 3) focused on this topic [55, 56, 57, 58, 59]The most suitable values for accelerating the card were obtained from these studies, as shown in Table 5. *Table 5*. The exposed data show the normal values of the RPi 3 and obtained after the application of the acceleration; the use of these data generated full stability in the stress tests performed to the embedded platform.

### 4) OpenCV in Raspbian

OpenCV, the computer vision library of choice for the current development, was cloned and installed, making use of [60, 61] to achieve this we used a software that allows to compile libraries and is used extensively in Linux distributions, known as CMake [62].

## B. Segmentation of objects by color

Taking the code of the algorithm and the above mentioned, the structure of the implemented software was identified, as shown in *Figure 4*.

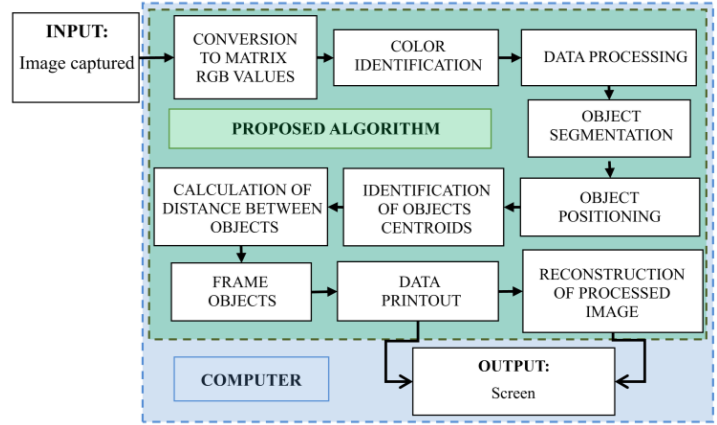

*Figure 4. Block diagram of the proposed algorithm.*

### 1) Algorithm analysis and results

Once the algorithm was developed, it was tested on a mid-range personal computer with the characteristics indicated in Table 6. *Table 6*The algorithm was then changed during the testing phase, due to the acquisition of a computer with higher performance.

*Table 6. Computer features used.*

| **Processor:** | Intel Pentium 2.2GHz |
|---|---|
| **RAM memory:** | 6 GB |
| **Video card:** | Integrated Intel(R) HD Graphics |
| **Camera:** | Logitech HD Pro Webcam C920 |
| **Operating system:** | Windows 7 Ultimate - 64bits |
| **NetBeans IDE:** | 8.1 |
| **Java(TM):** | 1.8.0_45-b15 |
| **OpenCV** | 3.1.0 |

In order to perform the tests, we took a count of 5 measurements or executions of the algorithm, in order to have a margin approaching the real average value of $fps$These measurements were made through a software designed in Java, which captured the start and end time and the number of frames reached to process in that period; as a starting point, the data was taken without the application of the algorithm, which only involves making the sample of images captured using OpenCV and Java, identifying the average frequency of $22.18 fps$ which leaves us a working margin of $7.18 fps$ to achieve the expected margin of $15 fps$ defined in III.D.4) it should be noted that this sampling is also due to the fact that the maximum transfer rate of the camera used in the implementation phase was max. $30 fps$. Next, the behavior of part of the algorithm was identified, so we proceeded to perform the test only with color detection, in order to observe the computational load that this task entails.

*Table 7. Results - Color identification only.*

| **Execution time (Sec)** | **No. Images processed** | **FPS** | **Sampling (Sec)** |
|---|---|---|---|
| 2 | 43 | 21,5 | 0,0465 |
| 2 | 44 | 22 | 0,0455 |
| 3 | 68 | 22,7 | 0,0441 |
| 6 | 102 | 17 | 0,0588 |
| 3 | 59 | 19,7 | 0,0508 |

This gives an average of $20,58 fps$ which represents a processing cost of $1.6 fps$which indicates that the cost of processing per image increased by 1.6 7,21%leaving a margin with respect to the expected value of $5.58 fps$. Now, with the inclusion of the segmentation method presented in III.D.8) was obtained.

*Table 8. Results - Identification by color and segmentation.*

| **Execution time (Sec)** | **No. Images processed** | **FPS** | **Sampling (Sec)** |
|---|---|---|---|

| | 3 | 46 | 15,3 | 0,0652 |
| --- | --- | --- | --- | --- |
| | 2 | 42 | 21 | 0,0476 |
| | 2 | 46 | 23 | 0,0435 |
| | 5 | 82 | 16,4 | 0,0610 |
| | 2 | 40 | 20 | 0,0500 |

With an average of $19,14 fps$ presenting a small reduction in comparison with what was previously endorsed, with only $-1,44 fps$ This is due to the fact that the calculations made depend on values found with the previous process.

*Table 9. Results - Complete algorithm.*

| Execution time (Sec) | No. Images processed | FPS | Sampling (Sec) |
| --- | --- | --- | --- |
| 4 | 66 | 16,5 | 0,0606 |
| 3 | 51 | 17 | 0,0588 |
| 4 | 65 | 16,2 | 0,0615 |
| 4 | 70 | 18 | 0,0571 |
| 7 | 141 | 20 | 0,0496 |

The analysis of the complete algorithm was performed with the inclusion of the positioning and the identification of the object area as shown in Figure 6. *Figure 6* obtaining an average of $17,54 fps$ leaving a positive part compared to the expected average performance of the algorithm. It is worth mentioning that the different variations of the sampling frequencies were due to the series of objects used, in addition to the influence of processes and programs of the computer in which the tests were developed; which generated changes in the sampling period, resulting from the different computational loads and hence why the capture of several samples was performed.

Finally, it was ratified that the algorithm is capable of identifying, positioning, measuring and segmenting the elements within a closed environment with the proposed characteristics, with great efficiency since in the 100% of the cases the detection, segmentation and location of the elements was carried out, in comparison with its accuracy that was lower with a 90.71% compared to the data directly compared to the data directly matched with a 89.08% y 92.34% for object midpoint and object distance measurements respectively.

*C. Optimal Route Generation*

Based on the object segmentation algorithm, system modeling and final considerations, we proceeded to the implementation of the algorithm that together with these guidelines allowed us to generate the optimal route for the mobile platform.

*1) Considerations and Graphical Interface*

For this instance of implementation, different colors were defined to determine the elements immersed in the study environment; as has been highlighted on numerous occasions, allowing both the algorithm and the reader its easy identification, assigned the color green for the mobile platform ($Nodo\ Fuente = 1$), red for the arrival area ($Nodo\ Sumidero = m$) and black for the different objects and obstacles. Now we proceeded to make the respective modifications to the graphic interface in order to have a greater interaction of the variables immersed in the object detection algorithm as indicated *in Figure 5*.

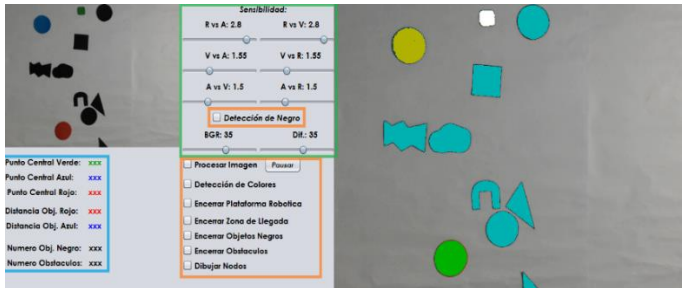

*Figure 5. Graphical User Interface implemented.*

Where the area framed in the green box represents the controls of the variables involved in the detection of the color of the objects, the orange area represents the indicators that allow to graphically identify each process, the blue box is the data exposure area and finally two areas that show the captured and processed frames located on the left and right respectively. The processed image of the *Figure 5* only shows the identification of the colors of the objects immersed in the study environment which has the necessary characteristics for the algorithm.

Next, the identification of each of the elements designated at the beginning of the current section was confirmed, resulting in the following figure *Figure 6*.

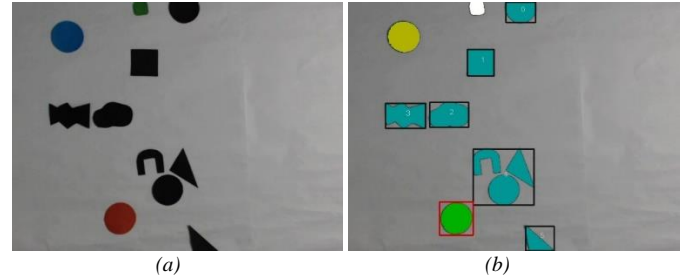

*Figure 6. (a) Captured image (b) Processed image.*

In which it is observed the full identification of the objects related above, which are framed with the colors of the detected object, thus indicating the initial node, final node, and possible obstacles, ignoring any other object not previously instantiated. It is also worth noting that the considerations set out in III.E.2).c) are taken into account, since although the possible very close obstacles have distances between them, they are not taken as distinct entities due to their reduced distance compared to the space necessary for the free movement of the robotic platform.

*2) Identification of real obstacles*

Although the data provided by the stage consisting of IV.B indicated the possible barriers between the moving platform and the finish area, they did not indicate the actual obstacles, since only objects that stand in the direct (diagonal or straight) path between the platform or a feasible node and the goal can be considered as an obstacle.

For this purpose, a method was used to define whether the possible obstacles are really one or not, through the use of the equation of the straight line, where using the limit points (corners) obtained from the mobile platform with respect to the midpoint of the arrival zone, a linear projection is generated, allowing to have an area where the platform would "cross" directly to the arrival zone, in case of identifying a direct intersection with any object that could be defined as an obstacle. The *Figure 7* is the result obtained by adding the indicated method.

Where it is identified that the real segmented and identified obstacles are framed in blue color, while the normal objects are in a black box, which indicates that the proposed method works properly, the green lines are the mathematical projection of the line from each corner to its middle position in the red zone, The green lines are the mathematical projection of the straight line from each corner to its middle position in the red zone, representing the green object (moving platform) in case it is positioned in that zone, which would be directly the imaginary projection of the moving platform on the center of the red object (arrival zone).

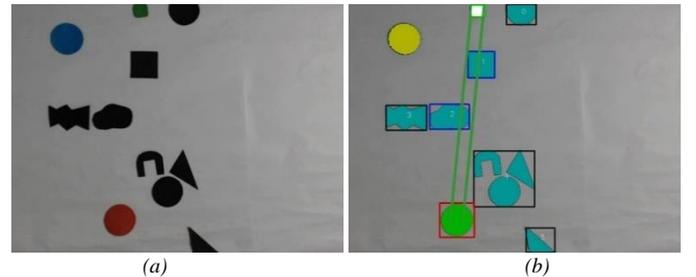

*Figure 7. Detección de obstáculos reales – (a) Imagen base y (b) Imagen procesada.*

We are aware of two start and end points $(P1(x_1, y_1))$ and end $(P2(x_2, y_2))$. In addition, we know that $A = x_2 - x_1 = y_u$  $y$ $B = y_2 - y_1 = -x_u$. And finally it is also known that $C = -By_1 - Ax_1$.

$$Ax + By + C = 0 \quad (9)$$

These considerations allowed us to calculate ranges of the passage area generated between the mobile platform and the target area, which allowed us to know if the obstacle is a direct barrier with the latter, therefore, if it is within the calculated area it would be an obstacle, but because the objects have different areas on the image, we used the set of ranges within which it could be an obstacle. For this we used the information of the segmentation algorithm to identify these ranges, for which we used the corners that make up the box that encloses the objects, thus generating 4 ranges per object corresponding to the sides of the area occupied by the object and evaluating the possible intersection of these ranges to identify whether it is an obstacle with respect to the arrival area. Finally, the event is iterated with all possible obstacles.

### 3) Identification of active nodes in the set *L*.

It should be recalled that the set *L* was used in III.E.2).a) and represents the set of all the nodes of the network, except for the source and sink nodes. Since the method used for the solution of the problem is based on the graph theory, it is unnecessary to include objects that do not represent a possible solution to the problem, which refers to the exclusion of objects that are not an impediment to the arrival zone, bringing with them a greater processing demand for the algorithm; therefore, we proceeded to create a programming instance which designates the participating nodes taking as a premise the: possible existence of barrier nodes (real obstacles), mandatory existence of a source node and a sink node.

This programming instance was performed based on the positioning of four active nodes per real obstacle identified, these nodes are represented in the corners of these obstacles, after the definition of these nodes along all the initial obstacles, we proceed to the verification of new obstacles with respect to the new nodes identified through the described in 1), this made it possible to verify the existence of all the nodes participating in the set. *L*. If there is no path to the sink node, it is considered as inappropriate and the platform does not perform any action.

#### a) Adjacency matrix generation

Following the idea previously stated, the code level generation of the adjacency matrix was carried out, which allows the abstraction of the network characteristics into a matrix of size $m \times m$ where ***m*** is defined by the number of nodes identified by the previous numeral, which are based on the information provided by the object segmentation algorithm implemented in B the obtaining of the guidelines used are shown in *Figure 8. Adyaceny matrix generation for the environment.* . Where the information of the segmented objects in are the input of the method which:

- *Define the possible feasible nodes (PNF):* through the location of four nodes per detected object; excluding the source and sink node that will only have its midpoint, these points located per object will be located at an angle of 45° with respect to its nearest corner, and distanced with a minimum distance, given by the size of the mobile platform, in order to ensure its passage in case of using the PNF.
- *Connection between NFP of the same object: it* was defined that, if a NFP belonged to the same object, it was necessary to identify whether it was possible to get from one node to another, by identifying obstructions on the box formed by the projection of the platform on the evaluated nodes. This would make it possible to ensure the passage of the platform from one NFP to another, if necessary.
- *Selection of feasible nodes (NF):* the information from the previous block was used for NF selection, taking as a premise that, if a PNF was isolated from the others (3 PNF), it was discarded as an inaccessible node, since it presents obstructions for its connection. This made it possible to discard unusable nodes, thus eliminating the algorithm load.
- *Identification of obstacles between NFs: the* detection of the obstacles between each NF was done using the method described in 2) was used, which would make it possible to define whether one NF was accessible from another NF. $NF_1$ was accessible from another NF $NF_2$ The platform could be moved from one NF to the other. $NF_1$ to $NF_2$ without any inconvenience. In such a way that the NF could be defined as a node participating in the network.
- *Calculation of distances between nodes:* making use of the equation (2) explained in III.D.7), the distance between nodes was calculated. Which we have defined as connection costs $c_{ij}$.
- *Addition of data to the adjacency matrix:* the connection data between the nodes evaluated in the immediately preceding phase are stored, allowing the complete abstraction of the network for its subsequent solution.

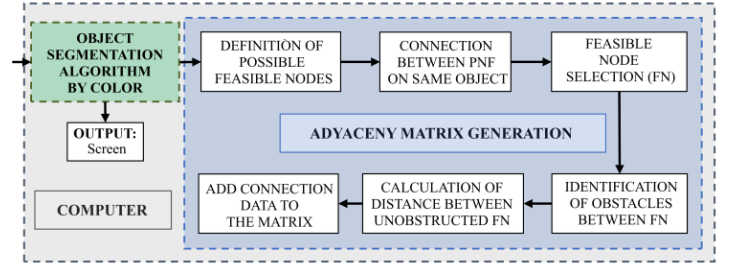

**Figure 8**. *Adyaceny matrix generation for the environment.*

It should be noted that the Adjacency Matrix will have a maximum size of $m_{max} = (n \times 4) + 2$ where **n** is the number of segmented objects, the additional one is generated by the source and sink nodes, these being nodes 1 and 2. 2 is generated by the source and sink nodes, being nodes 1 and 2 respectively. $m_{max}$ respectively.

### 4) Dijkstra solution method

In the implementation of the Dijkstra shortest path algorithm, it was done by means of the arguments presented in III.E.2).b) resulting in what is shown in the block diagram in *Figure 9*, which exposes the process performed by the Dijkstra algorithm which allows obtaining the optimal path of the adjacency matrix generated.

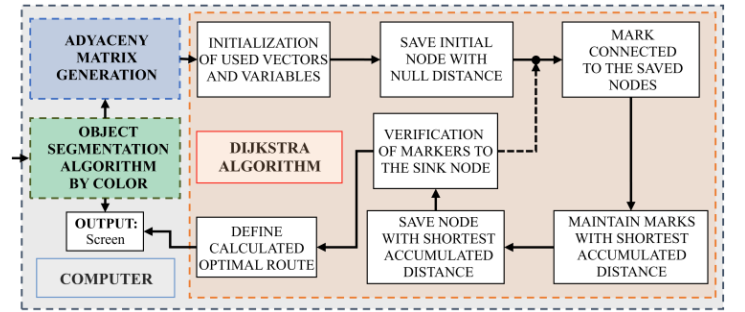

**Figure 9**. *Dijkstra algorithm implemented.*

- *Algorithm Initialization:* in this area all the variables, vectors and matrices to be used are arranged, of which it is important to mention the number of nodes, due to the above mentioned. $m_{max} = (n \times 4) + 2$ due to what is stated in a), matrix of saved nodes ($G \in \mathbb{R}^{3 \times m_{max}}$) and matrix of marked nodes ($M \in \mathbb{R}^{3 \times m_{max}}$); which include the distance traveled or accumulated, the node of origin and the number of iteration of the algorithm, finally the optimal route vector (), which contained the values obtained from the nodes and the number of the iteration of the algorithm. $R \in \mathbb{R}^{(m_{max}+1)}$ which contained the values obtained by the algorithm and in its last item stored the number of hops made to get from the final node to the initial one.
- *Save initial node: since* it is the node from which the algorithm must start, it is marked with a null distance value and is saved for its application within the algorithm, which allows to give the starting point to the algorithm.

$$G = \begin{bmatrix} 0 & inf & ... & inf \\ -1 & -1 & ... & -1 \\ 0 & 0 & ... & 0 \end{bmatrix}.$$

- *Mark nodes connected to the saved ones:* here the temporary marking of the nodes under evaluation was carried out, which, as already indicated, are placed in the matrix. *M*The data that could be transferred to the matrix are thus contained in the matrix. *G*.
- *Maintain marks with smaller distance:* it was identified if the node has old marks, if so, it is identified which is the smallest and it is left this one, which ensures to contain the predecessor node closer to the source node.
- *Save the node with the shortest distance:* the information is transferred from the node marked with the shortest distance to the matrix *G* and deleting it from the matrix *M*If the marked node is already in the matrix, the one with the shortest distance is identified and this matrix is updated with this label or mark. *G*If the marked node is already in the matrix, the one with the shortest distance is identified and the matrix is updated with this label or mark.
- *Verification of marks to the final node:* the verification of arrival labels to the final node was performed, where, if a final path is found, this is taken as a comparison entity with respect to the nodes of *M*If the distance value of the current tag from the final node in *G*is less than or equal to all the distances in *M*then the iterations are terminated because there is no way to improve the current distance. Otherwise, if there is still a possibility of improvement, continue iterating from the point of the current list. 3*er* point of the current list.
- *Define the optimal route:* for this purpose, use is made of the matrix *G*The optimal route: to do so, we use the matrix, on which we start to traverse from the final node, thus allowing us to reconstruct the shortest route from the tracking of the saved labels, using the space allocated for the predecessor node, and so on until we reach the source node.

a) Application of the method

The application of the method left a positive balance, indicating the optimal routes for the following scenarios individually. The images shown below indicate the effectiveness of the algorithm which, even with deplorable lighting conditions and an environment full of unnecessary objects, generates the optimal route adequately.

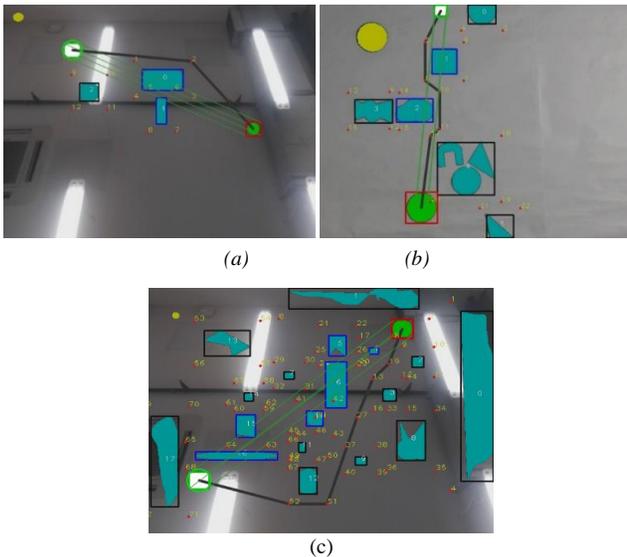

*Figure 10. Results of the Dijkstra method - a) 3 objects, b) 6 objects and c) 18 objects.*

It should be noted that this was achieved thanks to the robustness of the segmentation algorithm and the provision of controls in the graphical interface, which allowed the adequacy of the segmentation ranges. It is worth mentioning that these images were generated as an example for the current section, in the following section you can see the tests carried out to scale in an environment much closer to the real one.

### D. Data Acquisition Software

Making use of the hardware and software base provided in the previous paragraph (IV.A), ), on the RPi 3 embedded platform, a software dedicated to data acquisition was created and implemented, which allowed capturing images and sending them to subsequent stages, making use of all the constructs evaluated in the design stage. Until that moment, the implementation and data acquisition was not done through this platform, but it directly followed the characteristics required by the design. Now, with the inclusion of this module, the design implementation cycle is closed and allows the development of the complete system testing.

The software was implemented in JAVA language and designed through the NETBEANS 8.1 programming environment as specified in II.C, which allowed capturing images through the PI NOIR CAMERA V2.1 module, converting them into image files (.jpg) and sending them to the computer in charge of processing the acquired information through TCP (Transmission Control Protocol).

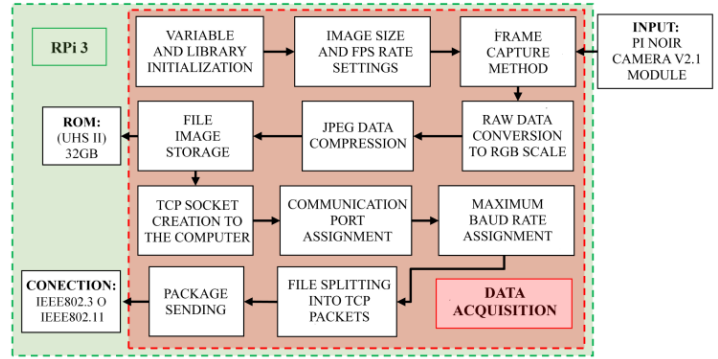

*Figure 11. Implemented software on Raspberry Pi 3.*

The implemented program follows the structure shown in *Figure 11*The program implemented follows the structure shown in Figure 11, which covered the tasks that fully fulfilled the purpose of the program. It is worth mentioning that packets were sent via TCP socket through the enabled port to the computer using either of the two communication media (Ethernet/IEEE802.3 [63] or Wifi/IEEE802.11 [64]) arranged on the platform, which the OS is in charge of managing the packet sending, the network connection and finally sends the data to the server with the previously assigned IP.

## V. EXPERIMENTS AND RESULTS

The current numeral reflects the tests performed and the additional considerations identified in the current phase, helping to successfully complete the development of the current project, it is here where the functionality of the implemented control was evaluated, and the system tests were developed as a whole.

### A. Modifications and Adjustments

Before the tests, we proceeded to make some modifications and adjustments that allowed the implemented system to perform efficiently, this was done following the methodological structure that governs the project, which allows the feedback of all the stages from any point of the project.

#### 1) Connection between systems

Due to the fact that in the implementation of the defined stages it was not necessary to connect them, the culmination of the connection is carried out in this instance in order to obtain the final system of the project, which is represented in the *Figure 12*.

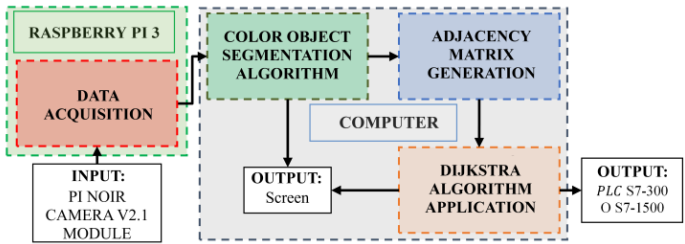
*Figure 12. Final block diagram of project system.*

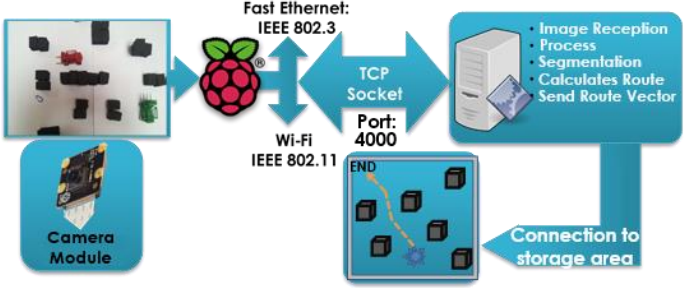
*Figure 13. General outline of development.*

In this case, the TCP protocol was used, which provides security in the transfer of information and is carried out by means of internationally endorsed standards (Ethernet/IEEE802.3 [63] or Wifi/IEEE802.11 [64]); by means of which each element is interconnected to form the final system; these elements are supported by at least one of these means of communication.

The general scheme of the *Figure 13* exposes the system process, where the capture of images is performed through a camera module, using the software developed and mounted on the RPi 3 platform, there average of a local area connection provided, communication with the server or computer responsible for performing the processing task, who finally generates the optimal route result of the system; which is sent through the same means of communication to the actuators of the plant, making use of the connection to the PLCs that support the basic control process.

### 2) Change of computer

In search of a higher performance, the acquisition of equipment with better performance compared to those presented in the implementation phase was carried out (*Table 6*). Among the major advantages are the use of a lighter and faster OS, a more recent processor, with better processing speed and greater performance of the graphics card, as shown in *Table 10*.

*Table 10. Software and hardware features of the new equipment.*

| Processor: | Intel Core i7 3.1GHz |
|---|---|
| RAM memory: | 6 GB |
| Video card: | NVIDIA NVS 3100M - 512MB |
| Camera | RPi 3 - Acquisition software |
| Operating system: | Ubuntu 16.0.4 LTS |
| NetBeans IDE: | 8.2 |
| Java(TM): | 1.8.0_111 |
| OpenCV | 3.1.0 |

This allowed the processing tasks to be carried out with a better performance rate compared to the previous development in implementation going from $17,54 fps$ a $29,3 fps$ in the segmentation of objects improving a 40,13% of the sequential type algorithm.

### 3) Parallelism of algorithms and processes

Since the execution of the processes and algorithms that take place within each phase of the project did not depend directly on a sequential interaction, the selection and parallelization of processes that would take advantage of the resources offered by the RPi 3 equipment and the computer was carried out. Where, for example, there are 4 cores inside the processor offered by the RPi 3, which allowed in theory to run four threads at the maximum speed achieved after overcloking ($1,4GHz$).

This allowed the implemented algorithms to run efficiently, speeding them up and improving their efficiency, having a positive and direct impact on the amount of $FPS$ In terms of control, this increased the level of sampling and allowed to generate a faster reaction to dynamic events.

Lo descrito permitió que los algoritmos implementados ejecutaran de forma eficaz acelerándolos y mejorando su eficacia, repercutiendo de forma positiva y directa sobre la cantidad de $FPS$, que el sistema pudo llegar a procesar, esto en términos de control aumentó el nivel de muestreo y permitió generar una reacción a eventos dinámicos más rápida.

To achieve this, we used one of the features of the selected language (JAVA), which allows the management of threads within the system, which are executed in different cores (parallelism). [65]), or otherwise in the same core but performing a time multiplexing for each process, which allows developing virtually parallel tasks, the latter is done when the hardware resources are not sufficient for the running software [66].

### B. Controlled Test Environment

For the development of tests, a scale assembly of the warehouse process area was carried out, which allowed the effectiveness of the control developed to be identified as closely as possible, therefore, the scale to be used on the environment was identified.

### 1) Development scale

The scale used allowed the design of the test system, for the definition of this we chose to search for a forklift model in the collectors market, where it used a vehicle with a ratio of 1 a 24 (1: 24), from this the whole scheme was developed.

With this in mind, the measurement of the basic element of the storage inventory (basket) was made, which contains 30 bottles, located in arrangements of $5 \times 6$ and using the information found in the documentation [15, 16, 20, 21, 22] it was identified that the palletizing system has 9 cases per level ($3 \times 3$) and have a range from 5 a 10 These data allowed the characterization of the obstacles immersed in the system.

*Table 11. Obstacle size.*

| | Basket | | Palletizing | |
|---|---|---|---|---|
| | *Real (cm)* | *Scale (cm)* | *Real (m)* | *Scale (cm)* |
| **Long** | 42 | 1,75 | 1,26 | 5,25 |
| **Width** | 35 | 1,458 | 1,05 | 4,375 |
| **High** | 24 | 1 | 1,2 - 2,4 | 5 - 10 |

Using the plans provided by an expansion project generated on the company Bavaria S.A., which are publicly available for viewing at [67](*due to copyright issues and possible claims, we will not include any plans*), these plans are arranged in the reference, from which we abstracted the data shown in *Table 12, the data shown in were abstracted from the reference,* forming the measurements used for the recreation of the test environment.

*Table 12. Storage and test area measurements.*

| | Warehousing | |
|---|---|---|
| | *Real (m)* | *Scale (cm)* |
| **Long** | 40 | 166,666 |
| **Width** | 32 | 133,333 |
| **High** | 15,5 | 64,5833 |

### C. Results Obtained

The current numeral contains the results obtained from different parts of the total system coupling, where the speed of capture, transmission, routing and total system solution were evaluated. All the measurements are focused on the number of $FPS$ The number of stages, which would later help to identify bottlenecks within the system, was identified. On the other hand, the possibility of scalability of the project was identified, going on to generate the optimal route for more forklifts at the same time. Finalizing with the data interconnection method between the system and the PLC.

*1) Capture speed*

The capture speed obtained by the software mounted on the embedded platform (RPi 3) was tested, which will give us the number of possible captures by the system. *FPS* possible, by the system. This would give us the sampling period generated by the acquisition software implemented.

*Table 13. Results obtained in the image capture*

| Execution time (ms) | No. Images Captured | FPS | Sampling (ms) |
| --- | --- | --- | --- |
| 632 | 57 | 90,18987 | 11,087719 |
| 3113 | 281 | 90,26662 | 11,078292 |
| 1486 | 134 | 90,17496 | 11,089552 |
| 1122 | 101 | 90,01782 | 11,108911 |
| 387 | 35 | 90,43928 | 11,057143 |

The definition of the execution time of the algorithm was performed randomly, where a certain time after the start, the element in execution was stopped, which finished its last iteration and delivered the results. There is a clear improvement compared to the implementation stage on a computer (IV.B.1), where the camera module in conjunction with the acquisition software delivers an average of $90,21 fps$ of 75.01% improvement compared to those of the previous phase. $22,54 fps$ of the previous phase.

*Table 14. Results - saving captured images.*

| Execution time (ms) | No. Images Saved | FPS | Sampling (ms) |
| --- | --- | --- | --- |
| 689 | 35 | 50,79826 | 19,685715 |
| 4849 | 211 | 43,51412 | 22,981043 |
| 3253 | 149 | 45,80387 | 21,832214 |
| 1204 | 63 | 52,32558 | 19,11111 |
| 1651 | 77 | 46,6384 | 21,441559 |

The above table reflects a loss in the speed of execution of the algorithm due to the encoding and conversion of raw data to RGB in JPEG compression, which allows the generation of image files. Leaving an average $47,81 fps$.

*2) Transfer of information*

The maximum number of images that could be sent via TCP was identified, which are restricted to the operating speed of the interconnected systems (RPi3 - Computer) and their means of transfer (Ethernet or Wifi), result on *Figure 15*.

*Table 15. Results in information transfer.*

| Execution time (ms) | No. Images Sent | FPS | Sampling (ms) |
| --- | --- | --- | --- |
| 1118 | 175 | 156,52951 | 6,3885713 |
| 593 | 100 | 168,63406 | 5,93 |
| 204 | 39 | 191,17647 | 5,230769 |
| 854 | 138 | 161,5925 | 6,188406 |
| 1423 | 227 | 159,52214 | 6,2687225 |

This made it possible to identify that the information transfer capacity allowed pass the $47,81 fps$ average, identified for the capture and storage, which are easily covered by the $167,49 fps$ the communication channel.

*3) Segmentation algorithm, adjacency matrix and optimal path generation*

For this case we evaluate the effectiveness of the set of algorithms described and implemented in the previous paragraphs B and C, where, thanks to the application of parallelism and process management, it was possible to improve even on the results of 2), This is reflected in the results shown in *Table 16*.

The average obtained from $50,31 fps$ outperforms per 21,01 frames to the algorithm applied without parallelization, providing an improvement in the processing speed of the system. 41,76%. It should be noted that this processing level does not exceed the communication speed previously evaluated, which revealed that this speed was sufficient to feed the necessary sampling of the system; however, the acquisition speed would represent a loss of speed in the overall execution, as it is lower.

*Table 16. Results of computer algorithm execution.*

| Execution time (ms) | No. Images Solved | FPS | Sampling (ms) |
| --- | --- | --- | --- |
| 4710 | 249 | 52,86624 | 18,915663 |
| 2433 | 123 | 50,55487 | 19,780487 |
| 938 | 48 | 51,172707 | 19,541666 |
| 2204 | 101 | 45,82577 | 21,821783 |
| 2190 | 112 | 51,141552 | 19,553572 |

*4) Total system execution*

Tests were carried out to verify the behavior of the system as a whole, which were based on tests performed on the environment described in B, allowing to evaluate the effectiveness of the implemented and coupled solution, where the segmentation and solution of the route was successfully performed in all the imposed cases, as long as they had any possible route.

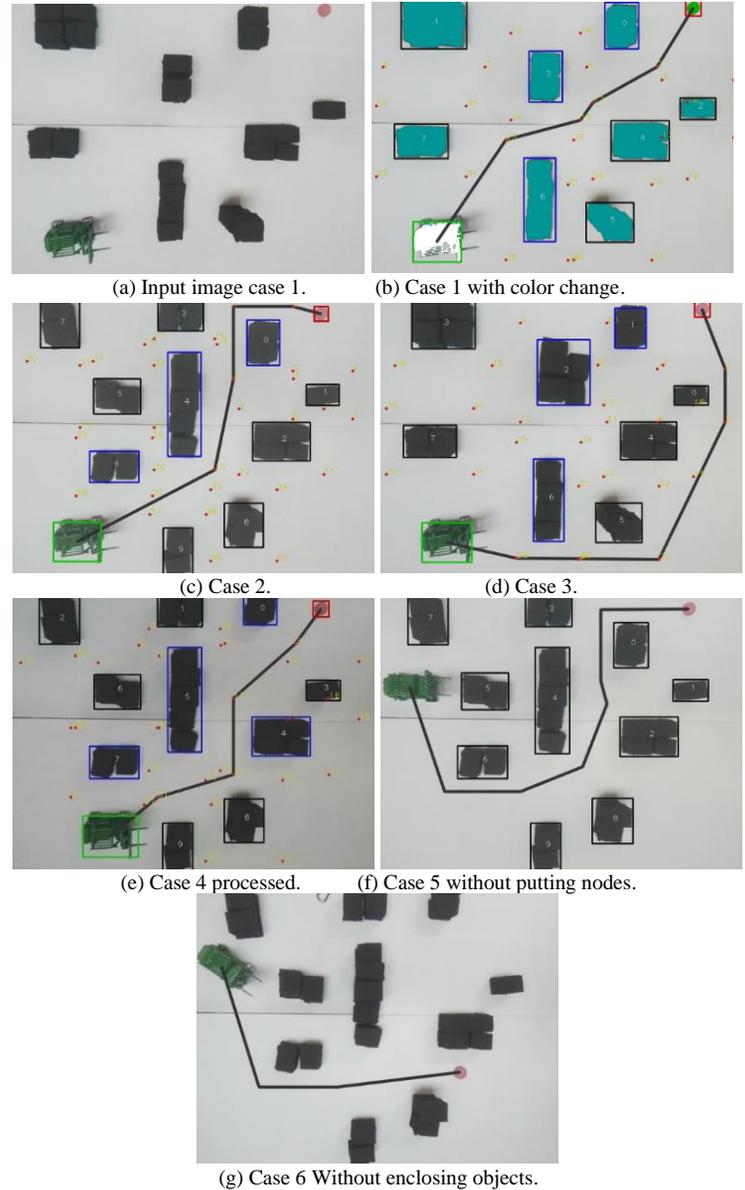

(a) Input image case 1. (b) Case 1 with color change.
(c) Case 2. (d) Case 3.
(e) Case 4 processed. (f) Case 5 without putting nodes.
(g) Case 6 Without enclosing objects.

*Figure 14. Test results - Small changes in the environment*

The cases shown above show some features included in the system such as color change, node display, object framing, among others. Small changes in the environment allowed to appreciate the response of the algorithm, which adapt to the conditions of the environment, depending for example on the passage of the green forklift, if it is no longer possible between two objects or if new spaces were generated.

*Table 17. System performance results*

| Execution time (ms) | No. Images Complete | FPS | Sampling (ms) |
|---|---|---|---|
| 6098 | 266 | 43,62086 | 22,924812 |
| 3836 | 144 | 37,539104 | 26,63889 |
| 6304 | 239 | 37,912437 | 26,37657 |
| 4500 | 187 | 41,555557 | 24,06417 |
| 2932 | 114 | 38,88131 | 25,719297 |

Now the result of the execution in terms of processing speed is reflected in *Table 17*, which has as average result $39,901 fps$ which correlates directly with the slowest process within the system chain shown in C.1), which corresponds to the acquisition of data by the RPi, and therefore the maximum number of FPS of the complete system was subject to the value of $47,81 fps$.

*5) System scalability*

The implemented system indicates the optimal routing of the test environment considering different selection criteria, therefore, for the generation of routing for several elements it is possible with small adjustments to it.

a) Change in the environment setup

The changes of the environment refer purely to the detection of the mobile platforms, which must have the necessary characteristics for the segmentation by color through the designed algorithm, therefore the colors were defined to perform it, eliminating the mark of the arrival position by color, in order to avoid inconveniences with the mobile platforms, therefore the arrival positions will be defined by the user through the graphical interface created, allowing to position the arrival site of the mobile platform in a freeway.

b) Changes in the adjacency matrix

The fundamental basis for the resolution of each case is the adjacency matrix, since it represented the abstracted graph of the test environment, it is necessary to identify the changes to which it was subjected for the resolution of more than one platform, therefore, it was identified that these new nodes could be added to the matrix that was previously defined as follows which would bring about the change represented below.

$$m_{max} = (n \times 4) + 2 \times p \quad (10)$$

Where $p$ represents the number of new mobile platforms, $n$ the number of objects within the environment and $m_{max}$ the maximum size of the adjacency matrix, which is reduced depending on the criteria set forth in IV.C.3).

Looking for the algorithm to have a higher performance, it was decided to use the adjacency matrix base, which we defined as the reference to all the elements that are feasible nodes except the arrival and departure nodes (in a few words the set $L$), which allowed us to avoid unnecessary and repetitive processing; thus solving this matrix for each specific case, which would only change the 4 vectors formed by the initial and final row and column.

$$M_{i \in P} = \begin{bmatrix} -1 & a_i & -1 \\ a_i' & K & b_i' \\ -1 & b_i & -1 \end{bmatrix} \quad (11)$$

Where $P \in \mathbb{N}$ is the set consisting of $\{1, 2, \dots, p\}$ being $p$ the number of different mobile platforms immersed in the environment. $i$ represents the initial and final node pair of a specific mobile platform. $M_i \in \mathbb{R}^{m \times m}$ is the adjacency matrix of the pair. $i$.

$K \in \mathbb{R}^{q \times q}$ is the base matrix composed of the distances of the possible connections between the nodes of the test environment, $a_i \in \mathbb{R}^q$ is the vector composed of the distances of possible connections between the partner's mobile platform and the nodes of the test environment. $i$ and the nodes of the test environment, $b_i \in \mathbb{R}^q$ is the vector composed of the distances of possible connections between the target point of the **pair and nodes of the test environment.** $i$ and the nodes of the test environment, $q$ is equal to the number of abstracted nodes of the system ($q = n \times 4$).

With this we would have an adjacency matrix for each platform, allowing the resolution linear problems individually and thus obtain answer, the process has advantages in that it uses a base matrix, which avoids calculating all the matrices and only focus on the resolution of the vectors $a_i$ y $b_i$.

a) Scalability results

The results obtained leave a positive side, where the resolution of environments with various mobile platforms can be observed, as illustrated in *Figure 15* and la *Figure 16*, which demonstrate the success obtained in terms of system scalability, allowing the generation of optimal routes for different forklifts within the test environment.

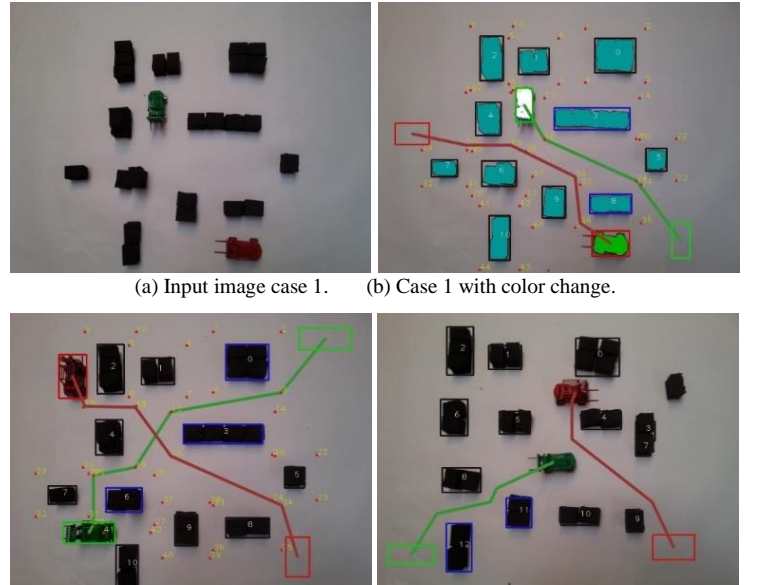

(a) Input image case 1.    (b) Case 1 with color change.

(c) Case 2.    (d) Case 3.
*Figure 15. Scalability - Two forklifts.*

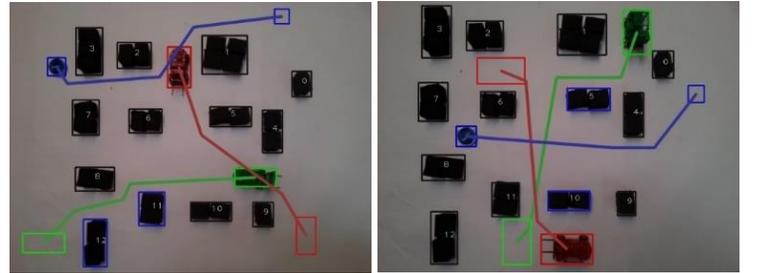

(c) Case 1.    (d) Case 2.

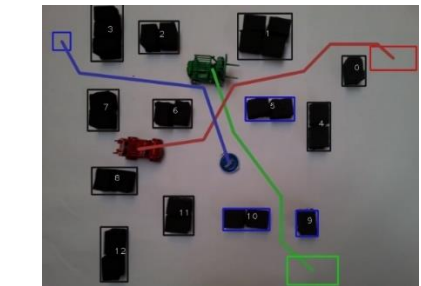

(c) ) Case 3 processed without enclosing platforms.
*Figure 16. Scalability - Three starting points.*

The solution of the exposed cases with two platforms is observed, demonstrating that regardless of the scenario, it generates the required route in an adequate way, where the arrival point is demarcated by a box with the color of the platform that will go to the place, it should be noted that the arrival position is arbitrarily defined from the graphical user interface.

The figure above shows that the scalability process is possible with several platforms, which will have a delay effect on the path generation algorithm, which was mitigated by b), The above figure shows that the scalability process is possible with several platforms, which will have a delay effect in the route generation algorithm, which was mitigated by b), but this does not indicate that it did not have a decay in terms of processing speed, where it went to having $50,31 fps$ obtained, to having $43,978 fps$ the speed of the system decreased from the speed obtained in $6,332 fps$ by the new mobile platforms included.

*6) PLC connection*

For the connection made to the PLCs that are the direct relationship with the storage area for the execution of the optimal route obtained, we made use of the connection libraries for the JAVA programming language, known as MOKA7 [68] these libraries have GNU license, which allows their operation and adaptation by the user who accesses them.

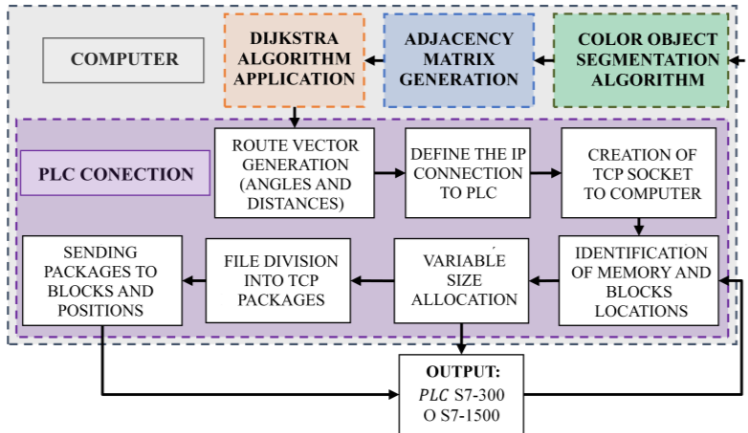

*Figure 17. Block diagram PLC connection.*

These libraries allow connection to Siemens S7300, S7400, S71200 and S71500 PLCs, a family of which are available, as previously described in (3), therefore, by using these libraries we had access to the following characteristics [69] such as:

- Connection by Ethernet or Wifi technologies.
- TCP connection with PLCs.
- Data verification.
- Development platform support (Eclipse and Netbeans)
- Connection to memory blocks.
- Modification and creation of PLC variables.
- Data transfer to memory blocks.

This allowed an adequate integration with the project, where the TCP protocol was used to connect to the equipment and send relevant variables for its control, as shown in the block diagram in *Figure 17*, as shown in the block diagram in Figure 17, the movement vector to be followed from the current position of the platform is sent, which is made up of the angle of rotation and the distance to be traveled for each section. This allows the PLC to obtain a tracking point and an input for its operation. But not before having made the identification of spaces and memory blocks used by the control.

## VI. CONCLUSIONS

The effectiveness of the object segmentation algorithm is remarkable due to its assertiveness, since it identified all the segmentable elements, immersed in the environments proposed for all the exposed cases, being this a good indicator of the level of robustness and reliability as a segmentation method. This effectiveness is far from its measurement precision, which presented an error of 9.29%, with respect to the data measured directly on the identified elements.

The final implemented system generates the optimal route in an adequate way, according to the constructs proposed throughout the project; where the average execution speed of the solution is notorious of $39,901 fps$ remembering that the current result was subordinated to the sampling period generated by the acquisition software exposed in section V.C.1), with an average execution period of $21,91 ms$ being the component or link of the system with the slowest execution speed.

The interconnectivity of the methods and modules of the system carried out by means of TCP, adequately uses the advantages immersed in the industrial environment under study, taking advantage of the resources offered by the industry, such as the local area network and the dedication of exclusive bandwidth of the same. This allowed to communicate the elements such as the PLC, the RPi and the server, in a reliable way and with a good data transmission rate. This allowed to affirm that the average data transfer speed obtained in the item "Data transfer rate" would not be mostly affected $167,49 fps$ average obtained in the item V.C.2) in a possible implementation in the real environment.

It was possible to eliminate the security scheme formed by the exclusive use of transportation routes, which limited the efficiency in the administration of the storage area, which is used because the processes immersed in this area are carried out by workers; this was improved through the implementation of the current system, allowing the movement of mobile platforms freely and autonomously within the area in question, thus helping to improve the execution of processes and administration.

The possibility of scalability or expansion of the system in terms of the number of mobile platforms, indicates the possibility of managing multiple sub-systems within the proposed environment, the cost of which is relegated to only $-6,332 fps$ by the inclusion of two additional platforms, indicating that the administration of a new element has a cost for the algorithm of $6,5\%$ with respect to the average of $fps$ of the inclusion of a previous element.

The included connectivity to the PLC modules, allowed to generate a level control approach, where the system could be controlled by PLCs as actuating elements, thus allowing to focus the development of the system more effective and less ambiguous in terms of optimal path generation; thus, ensuring that the system responds appropriately as proposed throughout the development.

## VII. REFERENCIAS